\documentclass[journal]{IEEEtran}
\usepackage{cite}

\ifCLASSINFOpdf
  \usepackage[pdftex]{graphicx}
\else
  \usepackage[dvips]{graphicx}
\fi
\usepackage[cmex10]{amsmath}
\usepackage{amssymb}
\usepackage{array}

\ifCLASSOPTIONcompsoc
  \usepackage[caption=false,font=normalsize,labelfont=sf,textfont=sf]{subfig}
\else
  \usepackage[caption=false,font=footnotesize]{subfig}
\fi

\ifCLASSOPTIONcaptionsoff
  \usepackage[nomarkers]{endfloat}
 \let\MYoriglatexcaption\caption
 \renewcommand{\caption}[2][\relax]{\MYoriglatexcaption[#2]{#2}}
\fi
\usepackage{url}

\begin{document}
%

\title{Scalability in Neural Control of Musculoskeletal Robots}
%
%

\author{
    \IEEEauthorblockN{Christoph~Richter, S\"oren~Jentzsch, Rafael~Hostettler, Jes\'us~A.~Garrido, Eduardo~Ros, Alois~Knoll, Florian~R\"ohrbein, Patrick~van~der~Smagt, and~J\"org~Conradt}

\IEEEcompsocitemizethanks{
\IEEEcompsocthanksitem C. Richter and J. Conradt: Neuroscientific System Theory, Department of Electrical and Computer Engineering, Technische Universit\"at M\"unchen, Germany. Correspondence should be addressed to {\tt c.richter@tum.de}.%
\IEEEcompsocthanksitem C. Richter: Bernstein Center for Computational Neuroscience Munich, Germany%
\IEEEcompsocthanksitem S. Jentzsch and P. van der Smagt: fortiss GmbH, Associate Institute of the Technische Universit\"at M\"unchen, Germany.%
\IEEEcompsocthanksitem R. Hostettler, F. R\"ohrbein, A. Knoll, and P. van der Smagt: Robotics and Embedded Systems, Department of Informatics, Technische Universit\"at M\"unchen, Germany.%
\IEEEcompsocthanksitem J.A. Garrido, E. Ros: Department of Computer Architecture and Technology, CITIC, University of Granada, Granada, Spain.%
}
}
\markboth{Richter \MakeLowercase{\textit{et al.}}: Scalability in Neural Control of Musculoskeletal Robots (IEEE R\MakeLowercase{obotics} \& A\MakeLowercase{utomation} M\MakeLowercase{agazine}, \MakeLowercase{accepted 2015-12-31})}{Richter \MakeLowercase{\textit{et al.}}: }

\maketitle

\begin{abstract}
Anthropomimetic robots are robots that sense, behave, interact and feel like humans. 
By this definition, anthropomimetic robots require human-like physical hardware and actuation, 
but also brain-like control and sensing. 
The most self-evident realization to meet those requirements would be a human-like musculoskeletal robot with a brain-like neural controller. 
While both musculoskeletal robotic hardware and neural control software have existed for decades, a scalable approach that could be used to build and control an anthropomimetic human-scale robot has not been demonstrated yet. 
Combining Myorobotics, a framework for musculoskeletal robot development, with SpiNNaker, a neuromorphic computing platform, we present the proof-of-principle of a system that can scale to dozens of neurally-controlled, physically compliant joints. 
At its core, it implements a closed-loop cerebellar model which provides real-time low-level neural control at minimal power consumption and maximal extensibility: 
higher-order (e.g., cortical) neural networks and neuromorphic sensors like silicon-retinae or -cochleae can naturally be incorporated. 
\end{abstract}

\begin{IEEEkeywords}
adaptive control, neurocontrollers, anthropomorphism, human-robot interaction, distributed processing, large-scale systems, parallel architectures, neural network hardware, biological neural networks, neurorobotics, learning (artificial intelligence)
\end{IEEEkeywords}

%
\section{Introduction}
\begin{figure}
	\centering
	\includegraphics[width=1.0\linewidth,clip,trim=0px 0 0px 0]{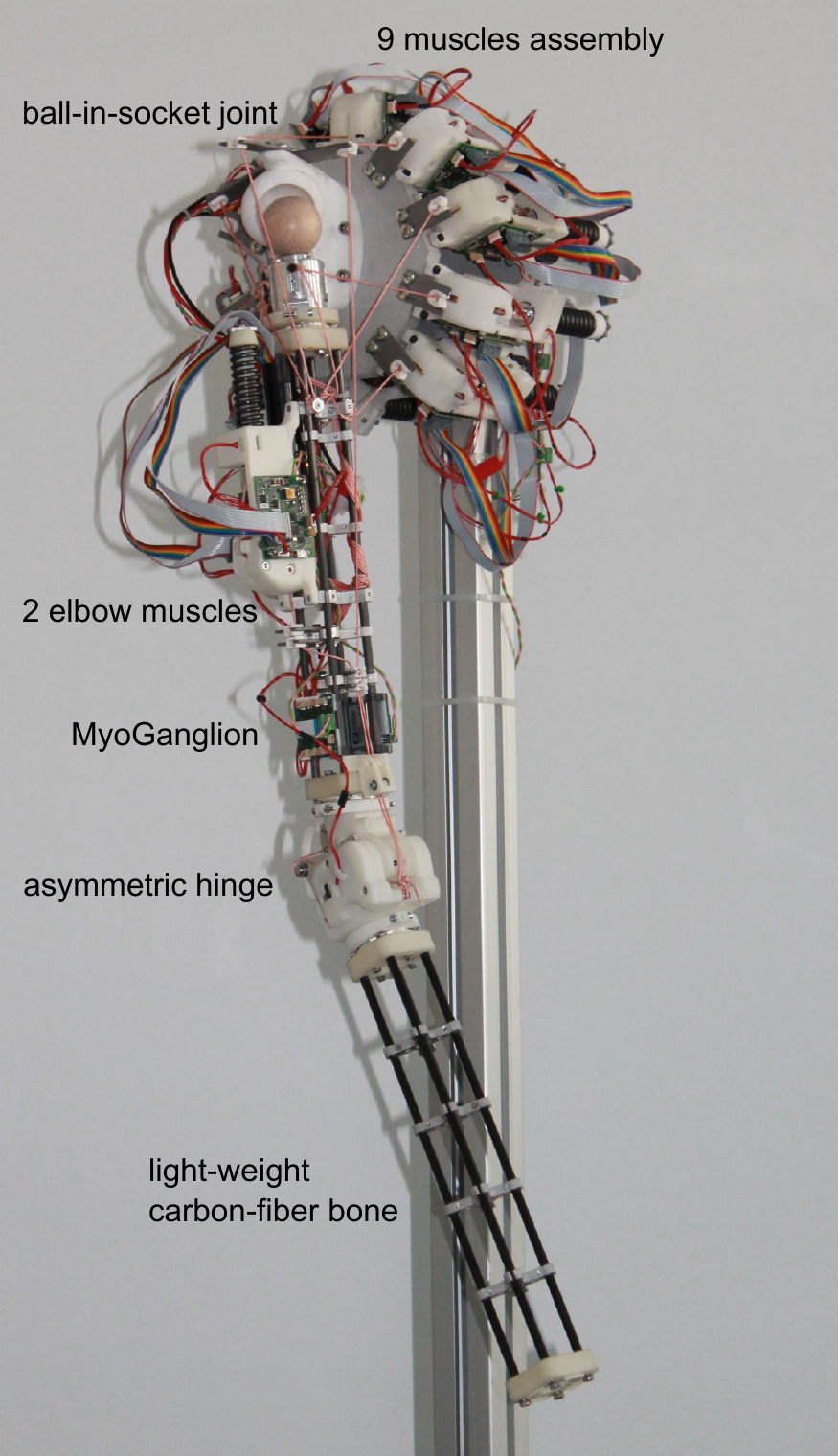}
	\caption{Complex Myorobotics arm mimicking the complexity of a human arm without spatula. 9 muscles cooperate to control the ball-in-socket-joint. One of these muscles, relating to the biceps, is biarticular, as it is attached so that it affects the motion of two joints, effectively coupling the shoulder and elbow joint.}
	\label{fig:anthropomimetic-robot}
\end{figure}

\IEEEPARstart{A}{}major challenge and vision for articulated robots is to behave and interact with humans in a safe and natural manner. Robots that mimic the mechanical properties of the human build strive toward both attributes simultaneously\cite{BicchiTonietti2004, Holland2006} as they possess built-in compliance and relatively natural, i.e., human-like, mass-distribution and dynamics by design. 
Musculoskeletal robots in particular offer lightweight, low-inertia end-effectors, since the main actuators, the skeletal muscles, can be kept at rest. 
Figure \ref{fig:anthropomimetic-robot} shows such a design, which coarsely mirrors a human arm. 
Most of the muscle mass is rigidly attached to the torso. 
Muscles connect to the distal bone only via tendons, which have a negligible weight. 
In this way, two passive safety aspects, which minimize the head injury criterion\cite{BicchiTonietti2004}, are intrinsic to the anthropomimetic musculoskeletal architecture: compliance and minimal moving mass. 

Similarly bio-inspired approaches on the controller side are simulated or emulated biological neural networks, because the (human) brain and central nervous system are the most relevant reference for \textit{natural} control of musculoskeletal limbs. 
Without doubt, neural control as done by animals or humans is the most elegant, versatile, and energy-efficient way to use musculoskeletal systems. 
Just as the human-like mechanical build has inherent passive safety advantages, brain-like control has desirable active safety features. 
The human nervous system implements active compliance on multiple levels. 
Arguably even more importantly, though, humans are perfectly accustomed to human-like behavior. 
Despite the fact that your colleagues could, if so inclined, injure you or others, working with humans is generally considered safe and does not require any special training. 
Consequently, there is every hope that their \textit{natural} and in this sense predictable behavior could gain anthropomimetic robots human-like safety attributes. 
\begin{figure}
	\centering
	\includegraphics[width=1.0\linewidth,clip,trim=0px 0 0px 0]{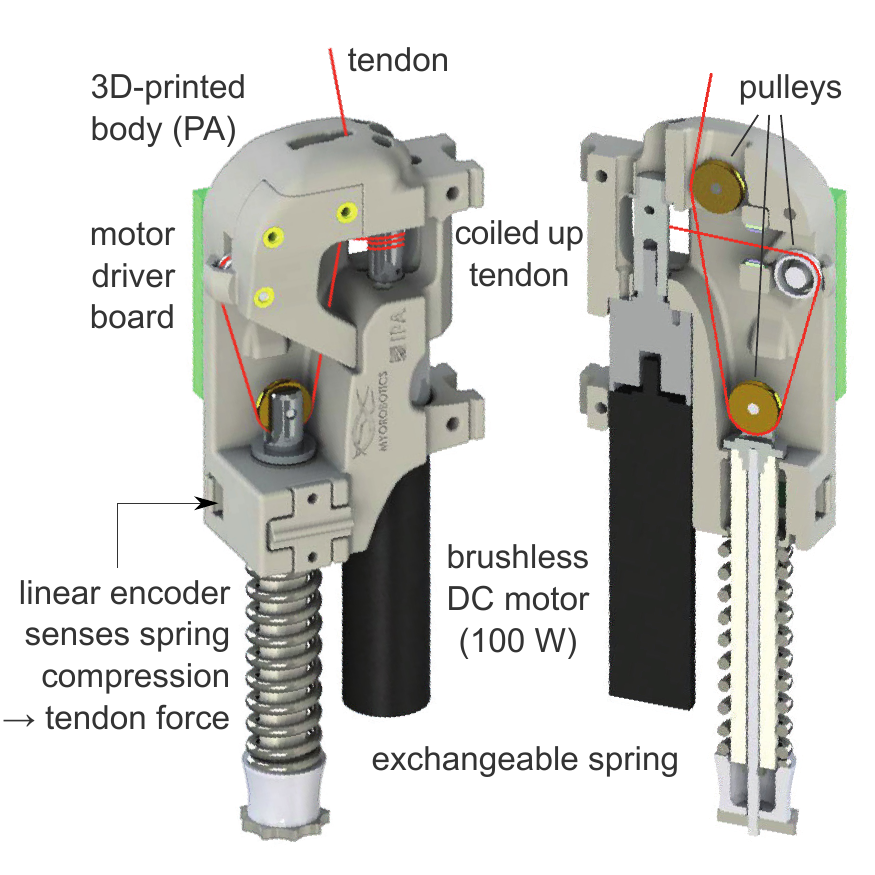}
	\caption{Myorobotics ``muscle'' with its components. The tendon (red cable) is routed in a triangular fashion in the muscle to create a non-linear net spring force. The tendon force is sensed by measuring the spring displacement through a magnetic strip fixed to the guiding rod of the spring that slides by a hall-effect encoder. This allows to calculate the respective force from a known spring constant and tendon routing geometry.}
	\label{fig:myo-muscle}
\end{figure}
The most demanding requirements and challenges on both the robotic hardware and the controller side are scalability and usability. 
Anthropomimetic robots have been built by numerous research groups, such as the Jouhou System Kougaku Laboratory of the University of Tokyo, and partners within the EU-funded project Embodied Cognition in a Compliantly Engineered Robot (Eccerobot)\cite{Pfeifer2013, Wittmeier2013a}, among others. However, those systems were custom-designed, mostly utilizing complex hardware and software, which inhibits (re-)production across labs and involves high production costs\cite{Marques2013}. 
The situation is similar with computing platforms. 
Robotic applications require flexible interfaces and strict real-time execution of large neural simulations\cite{conradt2000}. 
Different neuromorphic architectures and neuro accelerators have been developed during the previous decades, yet most of them, like those based on graphics processing units (GPU) \cite{Naveros2015,Yamazaki2013}, lack in terms of scalability. 
Special-purpose systems like those based on field programmable gate arrays (FPGA) \cite{Ros2006b,Agis2007} or custom silicon \cite{schemmel_wafer-scale_2010,Merolla2014} are usually too inflexible for a non-expert to implement and investigate custom learning rules, synapse types or cell models. 

To this end, the prevailing architecture for neural simulations and neural controllers is still the desktop computer, which we define in the context of this work as a Von Neumann architecture with a modest number of computing cores that share a common large random access memory. 
Depending on the underlying computations, such architectures are typically not optimal for simulating large neuronal networks\footnote{The human cerebellum alone comprises more than $10^{11}$ neurons\cite{Andersen1992}.}, which are inherently parallel\cite{Merolla2014}. 

In this article, we present the unique combination of musculoskeletal robotics hardware (Myorobotics) and neural control substrate implemented on a scalable spiking neural network infrastructure (SpiNNaker). 
We demonstrate how these technologies can address the aforementioned challenges and facilitate the development of human-scale anthropomimetic systems that are controlled by brain-like spiking neural networks. 
It is our conviction that SpiNNaker and Myorobotics pave the way for large-scale, complex neurorobots. 

\subsection{SpiNNaker}
SpiNNaker \cite{Furber2014} is a computer system designed for real-time simulations of spiking neural networks (SNNs) by the Manchester APT research group. 
A typical SpiNNaker system comprises thousands of ARM968 processing cores, which can run arbitrary code. 
They are distributed on a quasi-seamlessly extensible mesh network, which is spanned by special multicast routers at its nodes.
SpiNNaker's multicast routers are optimized for small (40 or 72\,bit-wide) data packets. 
Those SpiNNaker packets typically resemble action potentials or neural spikes in a SNN simulation. 
As such they typically convey only the source address of their originating neuron, from which the routers deduce the routing direction based on a user programmed routing table. 
Every SpiNNaker chip houses 1 router, 18 SpiNNaker cores (each with 96\,kB of local memory), and 128\,MB of shared SDRAM. 

The Manchester group provides an open software framework which promotes an event-driven programming model through the Spin1 API\cite{Furber2014}. 
Implementations of PyNN, a common interface for neuronal network simulators\cite{Davison2008}, and Nengo, a graphical and scripting based software package for simulating large-scale neural systems\cite{nengo2014}, are provided as a high-level, user friendly way to specify neural networks. 
These networks are then automatically mapped, uploaded, and executed on SpiNNaker. 
The entire software framework is open source, so it can be extended and modified by its users\footnote{https://github.com/SpiNNakerManchester}. 

In terms of SNN simulation performance SpiNNaker is superior to desktop computers by orders of magnitude. 
As a rule of thumb, a single SpiNNaker chip ($P \approx 1\,\mathrm{W}$) can handle a network of $10,000$ leaky integrate-and-fire neurons in real-time. 
A desktop computer needs a high-performance processor ($P \approx 50\,\mathrm{W}$) with fast memory to perform the same task. 
A single SpiNN-5 board contains 48 SpiNNaker chips drawing about the same amount of electrical power ($P \approx 50\,\mathrm{W}$), but providing about $50\times$ the computational power. 
Finally, the system scales from 18 (single chip) to over a million cores (65\,k chips), so the system size can be adapted to a wide range of neural network sizes, by interconnecting an appropriate number of SpiNNaker boards. 
Whereas the maximum system size might not be relevant to the robotics community, the scalability, power efficiency, flexibility and ease-of-use certainly are. 
SpiNNaker is designed for real-time SNN simulations. 
Given proper interfaces, it offers the prime opportunity to let large SNNs interact with and adapt to the real world. 

\subsection{Myorobotics}
Myorobotics is a toolkit for modular musculoskeletal robots that encompasses the full life-cycle of robot design. Robots can be assembled, optimized and simulated from primitives, then built and controlled from the same software. 
The robots are assembled from a set of primitives: ``bones'', ``muscles'', and joints, which are shown in Figure~\ref{fig:anthropomimetic-robot}. 
The most interesting of those building blocks, the ``muscle'', is detailed in Figure~\ref{fig:myo-muscle}. 
Its body is made of 3D-printed Polyamide (PA). 
It is actuated by a 100\,W DC motor (maxon EC) that coils up a cable, the ``tendon''. 
Three pulleys route the tendon in a triangular fashion. 
One of the pulleys is attached to a spring-loaded guiding rod. 
This mechanism endows the Myorobotics actuator with a (non-linear) series elasticity. 

The Myorobotics toolkit allows for the creation of a multitude of robot morphologies and enables researchers to investigate properties and dynamics of musculoskeletal robots. 
Its dedicated electronics provide tendon force, velocity, position and torque control at 500\,Hz directly from a standard desktop computer with all sensory data available on the bus. At this update rate the bandwidth of a single FlexRay interface, which is employed for high-level control, allows for up to 24 motors that can be driven concurrently. 

The framework can be easily extended with new primitives thanks to a standardized structure connector between the parts, as well as a software plugin that imports the construction directly from the CAD software SolidWorks. Consequentially, the system allows for primitives from a broad range of categories and covers many interesting use cases, such as anthropomimetic arms with complex shoulder joints (Figure~\ref{fig:anthropomimetic-robot}), quadrupeds and hopping robots. As the whole system was built with the non-robotic expert user in mind, it is easy to use and allows for fast modification of the robot topology. The entire system including all 3D models, schematics and all source code is open source\footnote{http://www.myorobotics.eu/}. 

What differentiates Myorobotics from other series elastic actuators and variable stiffness actuators is that Myorobotics actuators generate pulling forces between two attachment points rather than torques between two rigid links. 
This yields a fundamentally different control problem. 
While it can be reduced to classical joint-angle based control by describing a muscle Jacobian that maps the lengths of all tendons that apply forces between two links to a joint angle, this mapping is in many cases not unique; choosing a specific mapping means reducing the space of possible trajectories. 
However, it is currently subject of active research to design control strategies that directly map task space trajectories to desired muscle forces without the intermediate step of calculating target joint angles. 
This is especially interesting in the context of this paper, as all biological muscle based systems solve this control problem rather than a target joint angle/torque problem.  Myorobotics is thus a much closer model of the behavior of biological musculoskeletal systems than series elastic actuators like, e.g., MACCEPA. 

\subsection{Neural Circuitry}
We focus on implementing a cerebellar model to control the dynamics of the robotic system. 
Even though the major role of the cerebellum seems to be supervised learning of motor patterns\cite{doya2000complementary}, it is clear that vertebrate limb control can not be reduced to cerebellar functioning\cite{smagt2015neurorobotics}. 
An individual with cerebellar lesions may be able to move the arm to successfully reach a target, and to successfully adjust the hand to the size of an object. 
However, the action cannot be made swiftly and accurately, and the ability to coordinate the timing of the two subactions is lacking\cite{holmes1939cerebellum}. 
Vertebrate movement generation involves the basal ganglia, filtering out unwanted movements\cite{doya2000complementary}, as well as the motor and parietal cortices. 
Movement realisation, of course, also involves the spinal cord, which controls antagonism and seems to take care of nonlinearities in muscular functionality. 
Our model, however, focuses on a model of the cerebellar neurocircuitry for the following reasons. 
First, the fast learning of cerebellar circuitry is important for fast adaptation to environmental influences\cite{Luque2011b}. 
Second, some functionality of the spinal cord can be simulated with simple PID controllers\cite{bullock1991spinal}, especially for the comparatively simple actuator behavior that our system exhibits. 

\section{Setup}

\subsection{Robot}
\begin{figure}
	\centering
	\subfloat[]{
    	\includegraphics[height=2in,clip,trim=41pt 0 52pt 0]{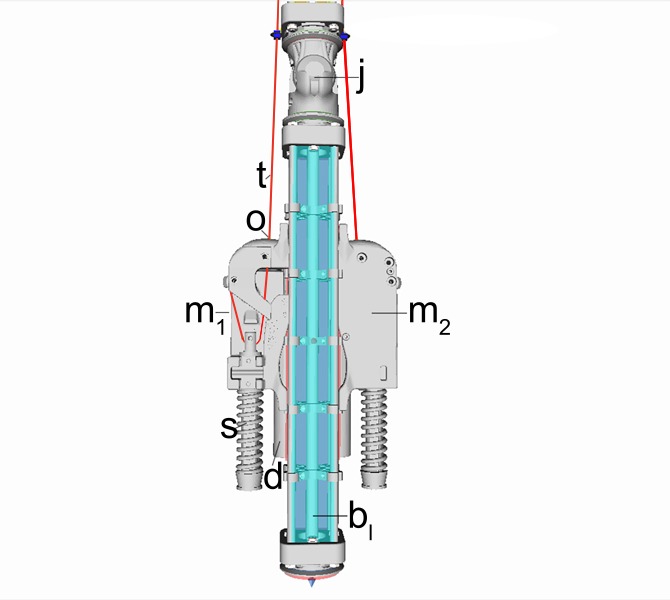}
    	\label{fig:single-joint-myo-schematic}
    	}
    \subfloat[]{
	\includegraphics[height=2in,clip,trim=4pt 0 0pt 0]{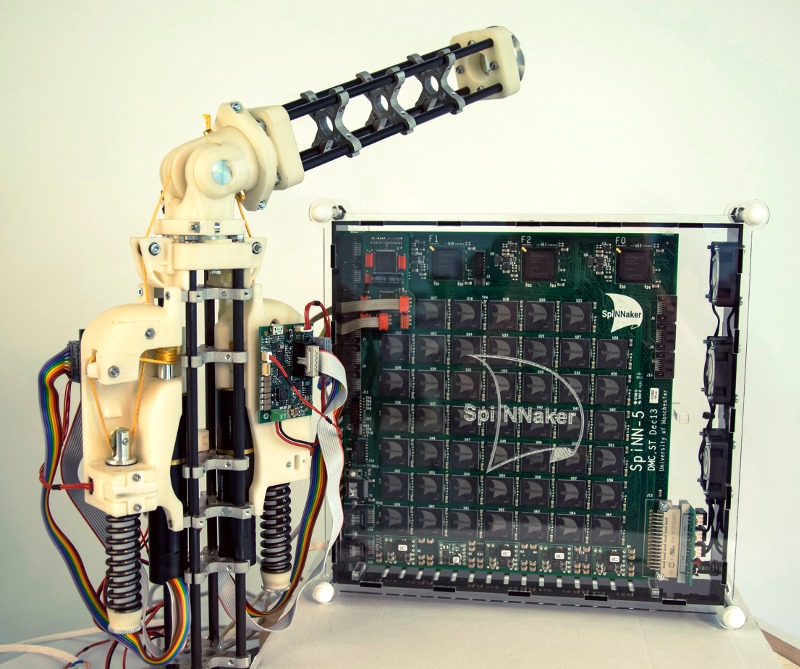}
	\label{fig:single-joint-photo}
    }
    \caption{Our single-joint Myorobotics proof-of-principle setup, as \protect\subref{fig:single-joint-myo-schematic} schematic (labels see main text), and as \protect\subref{fig:single-joint-photo} photograph with a SpiNN-5 48-chip SpiNNaker neuromorphic computer.}
\label{fig:single-joint+spinnaker}
\end{figure}

The robot employed in our proof-of-concept is the most basic setup that can be built with Myorobotics, consisting of a single symmetric hinge joint, two ``bones'' and two ``muscles'' driving it (Figure~\ref{fig:single-joint+spinnaker}). 
The system uses only the motor driver boards from the Myorobotics electronics, which we interface using a CAN bus.  
Larger Myorobotics systems connect the driver boards to intermediate controller boards (MyoGanglia), that offer a higher-level, higher-bandwidth control interface via FlexRay. 

Figure \ref{fig:single-joint-myo-schematic} highlights the individual parts of our joint assembly.
The two artificial ``muscles'' ($\mathrm{m_1}$, $\mathrm{m_{2}}$) are connected to the lower bone ($\mathrm{b_l}$), tendons connect them to the opposite side of the hinge joint (j).
Each ``muscle'' consists of a brushless DC motor (d) that coils up the tendon (t); we will call this \textsl{actuator} from now on.
The tendon is routed via a spring (s) and exits at a fixed outlet (o). 
The mechanically linear spring is combined with a triangular routing of the tendon (Figure~\ref{fig:myo-muscle}), making the net spring behavior non-linear. 
Since the actuators can only pull, an antagonist actuator is required. 
By pretensioning both actuators, both springs get contracted thereby changing the mechanical stiffness of the system. 
\subsection{Interfaces}
\begin{figure}
	\centering
	\includegraphics{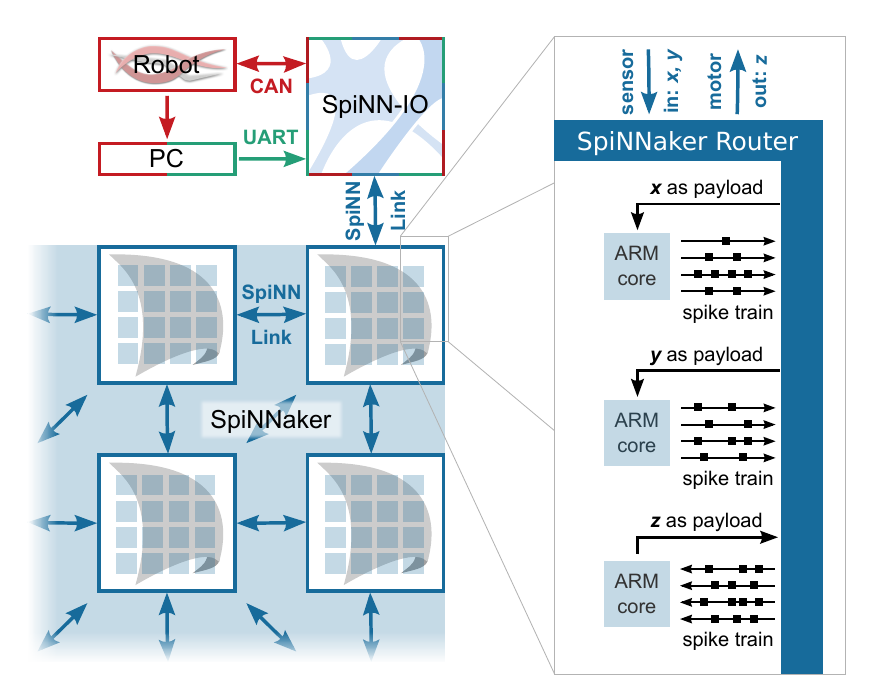}
	\caption{Logical and electrical layout of our system, frame and arrow colors indicate bus types. The SpiNN-IO board provides a real-time interface between the robot, the desktop computer, and SpiNNaker. By communicating with a SpiNNaker chip's router via SpiNNLink it provides the input SpiNNaker cores with sensory data $\mathbf{x,y}$. It receives motor commands $\mathbf{z}$ from output cores that it translates and forwards to the robot.}
	\label{fig:network-layout}
\end{figure}

To connect SpiNNaker to robotic sensors and actuators, we have developed a hardware interface that acts like another node on SpiNNaker's mesh network \cite{Denk2013a}. It translates sensor data into SpiNNaker packets and SpiNNaker packets into, e.g., motor commands. The microcontroller-based design allows us to connect SpiNNaker to many different bus systems including UART and CAN. We use the former for communication with an external desktop computer, the latter for Myorobotics actuators and sensors (Figure \ref{fig:network-layout}). Although the interface board allows for real-time injection of neural spike trains into SpiNNaker, our current implementation saves bandwidth by handling the de- and en-coding between robot data and neural spikes directly on SpiNNaker. 
The inset in Figure \ref{fig:network-layout} illustrates this setup: Sensory updates arrive as a SpiNNaker packet's payload at the respective ARM cores that are continuously emitting spike trains encoding the current sensory state. Likewise, dedicated motor cores continuously translate incoming spike trains to motor commands.

The typical translations performed on the input SpiNNaker cores are either rate- or population coding, 
the latter with Gaussian receptive fields and linearly distributed preferred values. 
Since SpiNNaker cores can be freely programmed, more flexible translation schemes, e.g. involving self-organizing maps, can be implemented. 
Our output cores translate the rate of incoming spikes from within the SpiNNaker mesh to a motor output signal via a linear transformation and an exponential fall-off in time. 
The time window and update cycle is typically 20\,ms. 
Again, more complex translation schemes can be implemented. 
Those could involve proprioceptive feedback from the Myorobotics actuators and in this way emulate the macroscopic or microscopic behavior of real skeletal muscles. 
It should be mentioned that from a PyNN network point of view, input and output are handled and set up like normal neural populations. 
The low-level implementation as C code is wrapped by PyNN objects and thus hidden from the PyNN programmer. 
All settings like time constants or scale factors can be adjusted in a user-friendly, object-oriented fashion. 

\subsection{Network Model}
\label{subsec:network}
As a first demonstration of our system we chose a cerebellar model that has previously been used to operate robots\cite{Luque2011b}. 
Our network model is akin to a Marr-Albus-style cerebellum \cite{Marr1969,Albus1971}. 
Its specific set-up including all cell parameters are derived from \cite{Luque2011b}. 
Although several network configurations were evaluated in \cite{Luque2011b}, we have considered the network that receives an implicit estimation of the robot actual state $\phi_\mathrm{act}$ and the set point $\phi_\mathrm{set}$. 
\begin{figure}
	\centering
	\includegraphics{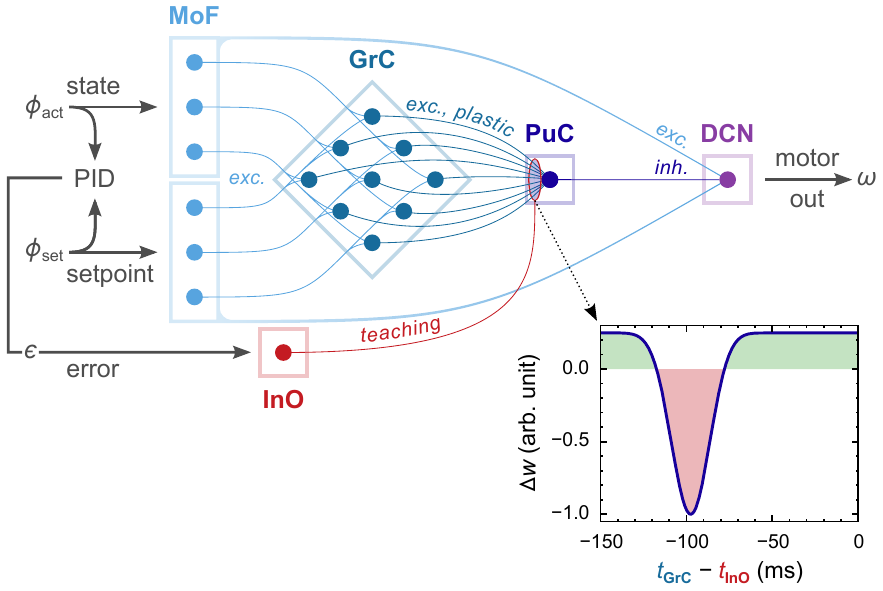}
	\caption{Schematic drawing of our cerebellar model along with its input (actual angle $\phi_\mathrm{act}$, setpoint $\phi_\mathrm{set}$, control error $\varepsilon$) and output (motor pulse width $\omega$). Single neurons or spike sources are represented by dots, populations are marked by rectangular frames, the thin lines connecting neurons represent synapses. The graph represents the learning rule governing weight changes $\Delta w$ at GrC-PuC synapses in response to the relative timing of GrC and InO spikes as they arrive at PuC dentrites.}
	\label{fig:cerebellum-model}
\end{figure}
Figure \ref{fig:cerebellum-model} illustrates the network structure. 
The network is comprised of leaky integrate-and-fire neurons with biologically realistic cell parameters and plausible divergence/convergence ratios between the different layers. As previously done in \cite{Luque2011b} we are omitting inhibitory interneurons and the olivo-cerebellar loop, to arrive at a most basic and deterministic model. 
However, the network still keeps the main roles that have been proposed for each layer in the Marr-Albus model \cite{Marr1969,Albus1971}, i.e., input sparse recoding of the mossy fiber (MoF) inputs in the granular layer and supervised learning in the Purkinje cells. 

Each of the two motors is controlled by the spike rate of 4 deep cerebellar nuclei (DCN) cells, which receive excitatory input from 32 MoFs, and inhibitory input from 8 Purkinje cells (PuC). 
MoF spiking activity (representing sensory input -actual state- and control data -target angle-) produces sequences of active granule cells (GrC). 
Since each of the 256 GrCs receives input from a unique set of MoF cells, a sparse coding of the input is made available in the parallel fibers (PFs), the long axons of the GrCs. 

The inhibitory corrective term that the DCN receives from PuCs is shaped through supervised learning between PFs and PuCs. The teaching signal encoding the actual error reaches the PuCs through the Inferior Olive (InO), producing complex spikes. 
This particular type of long-lasting spikes have been demonstrated to induce long-term depression in the PF-PuC synapses when correlated with simultaneous PF spikes \cite{Yang2014}. 
This learning mechanism has been implemented by using a kernel function $\Delta w(t_\mathrm{GrC} - t_\mathrm{InO})$ relating mutual InO--GrC spike timing with synaptic weight changes $\Delta w$ (see \cite{Luque2011b} for details). 
It basically punishes synapses that likely lead to erroneous behavior: 
If a GrC spike on a GrC-PuC synapse leads to some action and is followed by an InO spike after a characteristic response time, say 100\,ms, 
then the respective synaptic weight, which was likely responsible for that error, is depressed \cite{Luque2011a}. 
In order to compensate the long-term depression term, long-term potentiation is induced every time a presynaptic spike occurs in the PFs. 
The effective spike-timing dependent plasticity function $\Delta w(t_\mathrm{InO} - t_\mathrm{GrC})$ is plotted in Figure \ref{fig:cerebellum-model}. 
Interestingly, this learning rule also deals with the long delay that has been observed in the action-perception loop of the nervous system that has been estimated around 100\,ms \cite{Luque2011a}. 

This rather unusual learning rule would be impossible to implement on many neuro-accelerator platforms. SpiNNaker, on the other hand, is freely programmable. 
Just like the previously discussed input and output populations, we implemented this learning rule as low-level C code on SpiNNaker and wrapped it into a PyNN object for the high-level network description. We chose a look-up table (LUT) based approach, in which the LUTs for the temporal kernel are compiled by the Python frontend. The respective SpiNNaker cores buffer up to 160 spikes per simulated synapse and evaluate their mutual timing and the corresponding synaptic weight change periodically. 

The reference network implementation, which we ported to SpiNNaker, is running on EDLUT\footnote{Event-Driven simulator based on Look-Up-Tables, http://edlut.googlecode.com/} \cite{Ros2006a,Naveros2015}, 
a high-performance event-driven neural network simulator software. 
We developed a framework that translates a high-level text-based network description for either EDLUT or PyNN, runs the simulation on the PC or SpiNNaker, respectively, and compares the resulting network output. 
Through the use of a programmable power supply (Manson HCS-3202) we can monitor SpiNNaker's run-time power consumption and compare it to that of EDLUT running on our desktop computer. 
This way our SpiNNaker implementation could be rigorously checked and tested against the reference implementation on EDLUT. 

Our SpiNNaker implementation matches the EDLUT reference well. 
Minor deviations mainly stem from the fact that the SpiNNaker implementation is tick-based and uses fixed point representations while EDLUT is purely event-driven and uses double precision floating point. 
SpiNNaker cores lack a floating point unit, so floating point computations on SpiNNaker would be inefficient. 
The Manchester team made this design decision to save on power consumption and die area per core. 
A comparison to the relatively efficient desktop-based software simulator EDLUT highlights SpiNNaker's power efficiency: 
Depending on the network layout and its input, SpiNNaker's energy consumption is just one hundredth to one tenth that of EDLUT running on a typical desktop computer---a considerable asset that is especially relevant for autonomous robots. 
\subsection{Graphical User Interface}
While our system can run headless, in a closed-loop fashion, we have built a graphical user interface (GUI) for live monitoring and interaction with the neural simulation on SpiNNaker. The software runs on an external computing station, receives data from the CAN bus and uses UART to inject data into SpiNNaker via the SpiNN-IO board (Figure \ref{fig:network-layout}). 

Through the GUI, the user can control the target joint angle of the robot and adjust the PID parameters determining the teaching (error) signal calculation. 
The teaching signal as well the target angle are sent to SpiNNaker at an update frequency of 20\,Hz. 

As for debugging purposes and performance evaluation, the user can monitor the current CAN data, the deviation between current and target joint angle, the current error signal, as well as the live spike train of selected neuron populations. 
\section{Evaluation}
\begin{figure*}
    \centering
    \subfloat[naive cerebellum]{
        \label{fig:working-cerebellum:naive}
        \includegraphics[width=3.5in]{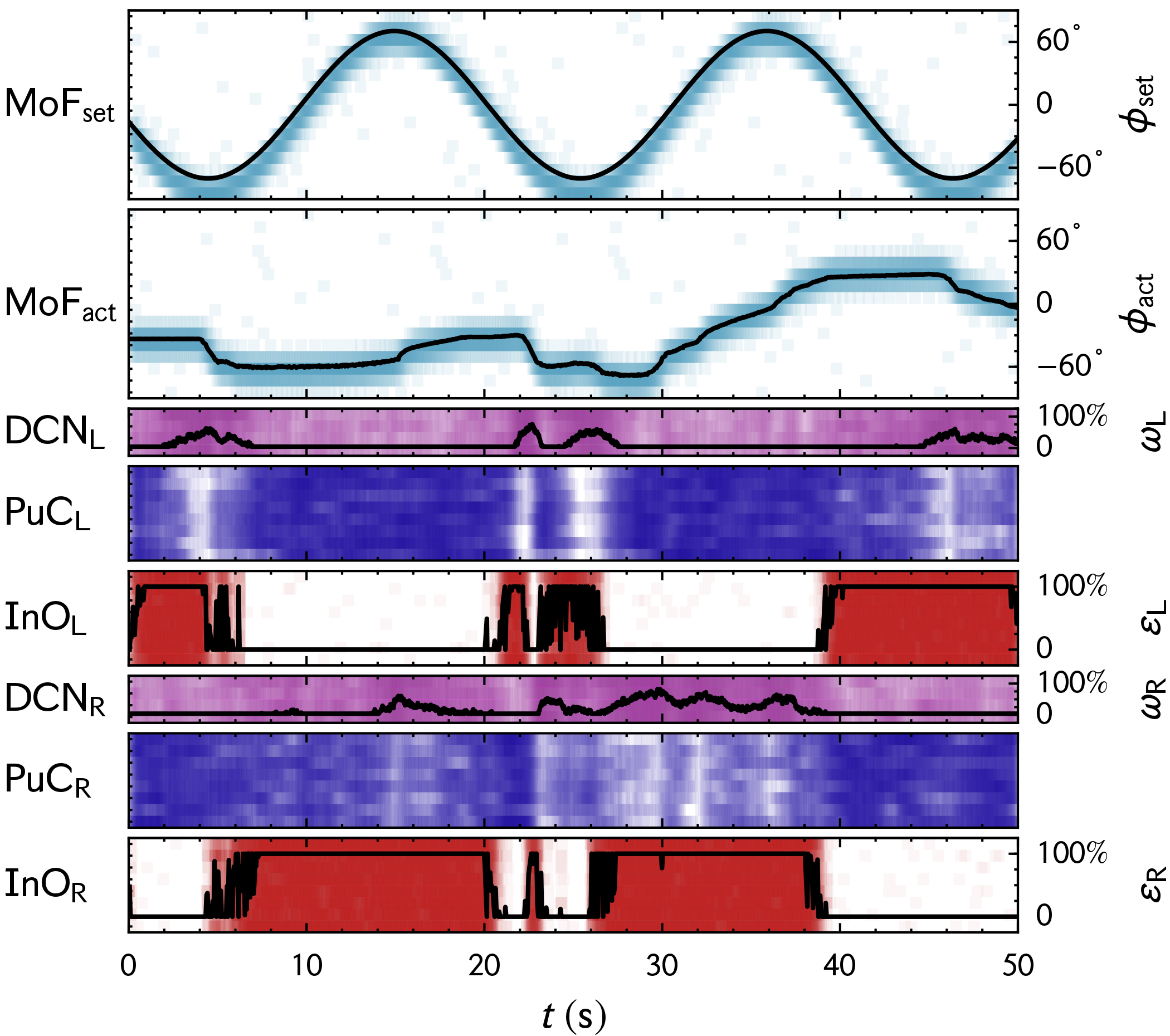}
    }
    \subfloat[five minutes later: trained cerebellum]{
        \label{fig:working-cerebellum:trained}
        \includegraphics[width=3.5in]{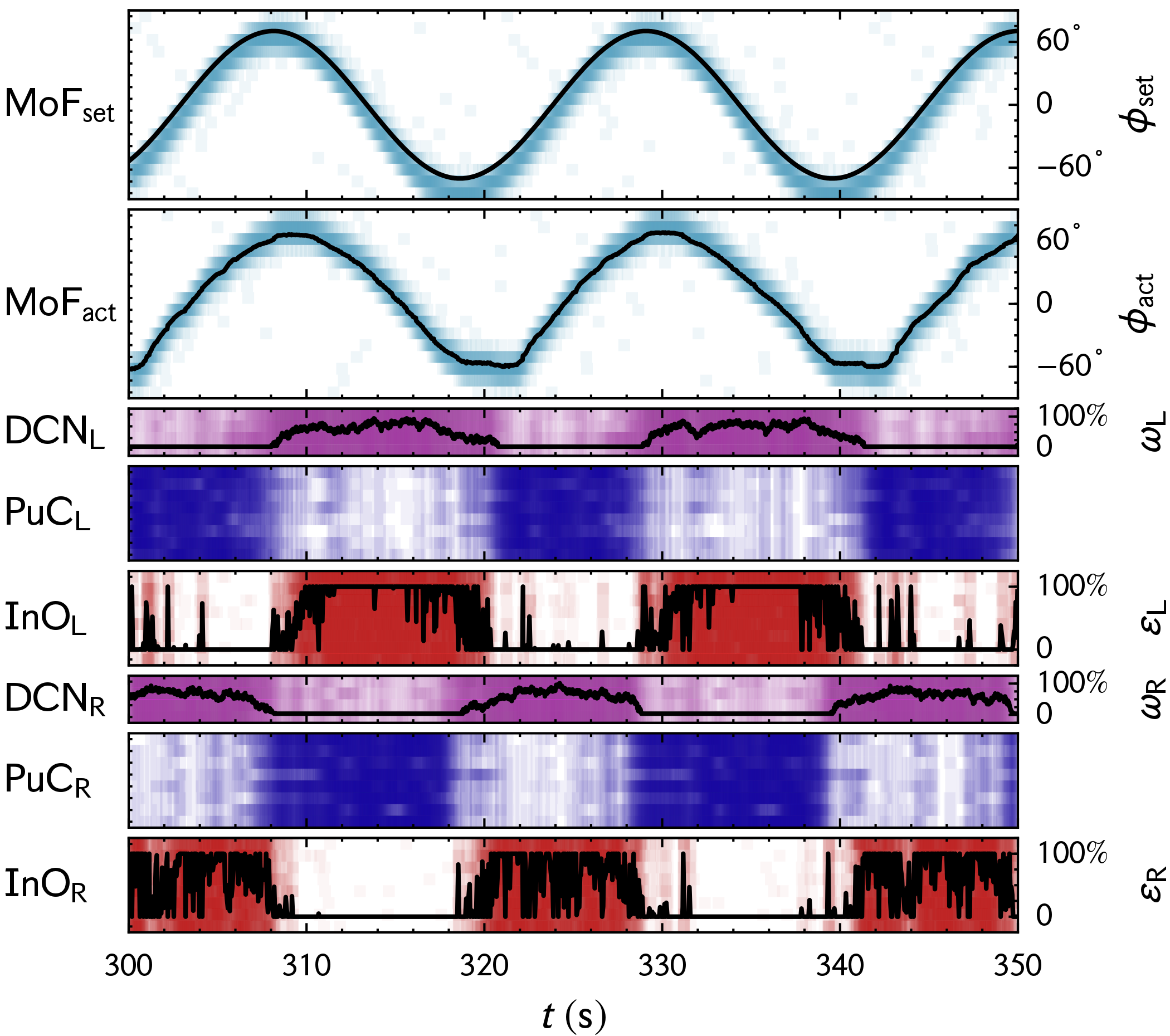}
    } 
\caption{Performance of our cerebellar real-time simulation on SpiNNaker. Colored semi-opaque squares symbolize spike events (spike density $\propto$ color saturation), left ordinate indicates the neuron ID. The black solid curves indicate the respective angle $\phi$ (MoF input), motor output $\omega$ (DCN output) or error signal $\epsilon$ (InO input). Indices: L, R = left, right; act: actual sensory values; set: setpoint. 
}
\label{fig:working-cerebellum}
\end{figure*}
Figure \ref{fig:working-cerebellum} shows the performance of our proof-of-principle cerebellar model running in real-time on SpiNNaker while controlling the antagonistic Myorobotics joint. 
It illustrates neural spikes as colored raster plots and its control and sensory input values over time as black solid lines, where applicable. 
There are separate PuC (dark blue spikes), InO (red spikes) and DCN (purple spikes) populations, for the left (index ``L'') and right (index ``R'') actuator. 
The error signals $\epsilon_\mathrm{L}, \epsilon_\mathrm{R}$ are computed as a PID error signal $E(\phi_\mathrm{set}-\phi_\mathrm{act})$ by the GUI. 
The corresponding spike trains are emitted by the InO populations.
They shape the weights between the GrCs (not shown) and the respective PuCs. 
The control input is the joint angle set point $\phi_\mathrm{set}(t)$ as given by the GUI, it corresponds closely to the mossy fiber spikes of the MoF$_\mathrm{set}$ population. 
The other control input is the actual joint angle $\phi_\mathrm{act}(t)$ as measured by a magnetic angle sensor within the Myorobotics joint, MoF$_\mathrm{act}$ is the corresponding neural population. 
The SpiNN-IO board reads $\phi_\mathrm{act}$ directly from the CAN bus, receives $\phi_\mathrm{set}, \epsilon_\mathrm{L}, \epsilon_\mathrm{R}$ via UART (from a USB-UART interface) and streams all values into SpiNNaker via SpiNN-Link where they are translated into spike trains as illustrated in Figure \ref{fig:network-layout}. 

\subsection{Control Performance}
In the present set up the teaching signal conveys the mismatch between the actual and the target joint angle. 
Consequently, in order to minimize that error and maximize the agreement between $\phi_\mathrm{act}$ and $\phi_\mathrm{set}$, the network learns to perform antagonistic control. 
In our scenario, the network learns to follow the given sinusoidal trajectory $\phi_\mathrm{set}(t)$ within few revolutions. 
The learning is exclusively based on the intrinsic plasticity mechanism at GrC-PuC synapses, as explained in subsection \ref{subsec:network}. 
Figure \ref{fig:working-cerebellum:naive} shows the performance of the naive network with randomly initialized weights at the GrC-PuC synapses. 
In this state the cerebellar model does not even know right from left. 
Therefore, the joint angle $\phi_\mathrm{act}$ (MoF$_\mathrm{act}$) does not track the given trajectory $\phi_\mathrm{set}(t)$ (MoF$_\mathrm{set}$) at all. 
As an example, at $t=15\,\mathrm{s}$, $\phi_\mathrm{set}$ is at the rightmost position, $\phi_\mathrm{act}$ still on the far left side. 
In this situation, the right muscle should clearly pull more.  
The teaching signal reacts accordingly: $\epsilon_\mathrm{R}$ is at its maximum resulting in a high InO$_\mathrm{R}$ firing rate. 
The corresponding GrC-PuC$_\mathrm{R}$ weights decrease accordingly. 
In subsequent similar situations this results in less DCN$_\mathrm{R}$ inhibition by the PuC$_\mathrm{R}$ population and more motor output $\omega_\mathrm{R}$. 
So after 5 minutes of run time (and learning) the system can follow the trajectory much better (Figure \ref{fig:working-cerebellum:trained}): 
$\phi_\mathrm{act}$ tracks $\phi_\mathrm{set}$ much more closely, the cerebellar model has learned to do antagonistic control. 
The cerebellum can also learn to follow different other waveforms or manually controlled trajectories\footnote{See \url{https://youtu.be/y6MwOtW3_kQ} for a video demonstration.}. 

Note that in the given example the network output is the sole control input to the robot arm. 
It controls the motors directly via pulse width modulation. 
While this nicely demonstrates the learning capabilities of the network, it does not mirror the biological antetype. 
In a more biologically realistic scenario, the cerebellum would output a corrective term that adds to a (cortical) forward-kinematic control signal. 

\subsection{Scalability and Constraints}
Our present configuration runs on a single SpiNNaker chip. 
It utilizes only 16 SpiNNaker cores, 2\,\% of a single SpiNN-5 board. 
Consequently, there is ample room for adding more joints and actuators as well as higher-level (e.g. cortical) neural networks. 
With SpiNNaker being a scalable system, computing resources are clearly not the bottleneck anymore. 

Our SpiNN-IO board connects the robot and the desktop computer with SpiNNaker. 
Its microcontroller limits the effective, combined update rate of input and output populations to about $500\,\mathrm{kHz}$\cite{Denk2013a}. 
In effect, our current system could handle 500 input/output populations at an update rate of 1\,kHz. 
As each input/output population occupies a single SpiNNaker core, those neural populations would fill about 60\,\% of a single SpiNN-5 board. 
A limit of 500 sensory streams at 1\,kHz update rate translates to roughly 100 actuators, or dozens of joints that could be controlled with a single or few SpiNN-5 boards---within an order of magnitude to a human-scale robot. 
This limit could be alleviated by using (a) more than one SpiNN-IO (on separate SpiNN-5 boards), or (b) a modified SpiNN-IO design that uses SpiNNaker's inter-board connectors. 
FPGAs on the SpiNN-5 board multiplex 8 SpiNNLink ports on each of those connectors. 

The remaining bottleneck is the communication between SpiNN-IO and the robot. 
In our current setting we can use up to 4 separate CAN busses, which can manage up to 4 joints (8 Myorobotics actuators) at an update rate of 500\,Hz. 
By using the full Myorobotics electronics, namely up to 6 MyoGanglia connected to a dedicated FlexRay controller, up to 12 joints (24 actuators) can be used at the same update rate. 
Again, with multiple SpiNN-IO boards, each with a dedicated FlexRay controller, we can alleviate this limit. 

\section{Discussion}

By demonstrating the control of a musculoskeletal joint with a simulated cerebellum running in real-time, we successfully combined robotic hardware (Myorobotics) and simulation platform (SpiNNaker). 
Both Myorobotics and SpiNNaker offer scalability and usability:
They can be extended in a straightforward manner, with no major roadblocks in sight towards systems approaching human-level complexity. 
Of course, many components still have to be added in order to arrive at a system that can interact with its environment in an intelligent way. 
Fortunately, a number of suitable technologies are readily available today. 

\subsection{Sensors}
While any sensor could be added to our framework, event-based systems are the most natural fit. 
Their sensory address-event-representation (AER) maps directly onto SpiNNaker packets, i.e. neural spikes in SNN simulations. 
The events emitted by an AER auditory sensor, or \textit{silicon cochlea}\cite{chan2007aer}, for instance, represent a sound's momentary frequency-resolved power spectrum. 
Their address encodes a specific frequency. 
The repetition rate of events with the same address encodes the respective spectral weight. 
Events emitted by an AER vision sensor, or \textit{silicon retina}, typically represent a sudden, pixel-local change in brightness. 
Here, the address encodes the pixel coordinate. 
Silicon retinae \cite{lichtsteiner2008,Denk2013a} and cochleae have previously been integrated with SpiNNaker. 
They are perfectly compatible with our interface. 

\subsection{Intelligence}
SpiNNaker can serve as a computing backend for PyNN\cite{Davison2008} or Nengo\cite{nengo2014}. 
Neural networks specified in those languages can often be run directly on SpiNNaker or require only minor modifications to be made, when porting a network from one compute backend to another. 
Missing software features, in our case a learning rule and input/output handlers, can be added to SpiNNaker's open source framework. 
Thus, many available models can be ported and integrated into the system with minor effort. 

\subsection{Systems}
\begin{figure}
	\centering
	\includegraphics[width=1.0\linewidth,clip,trim=0px 0 0px 0]{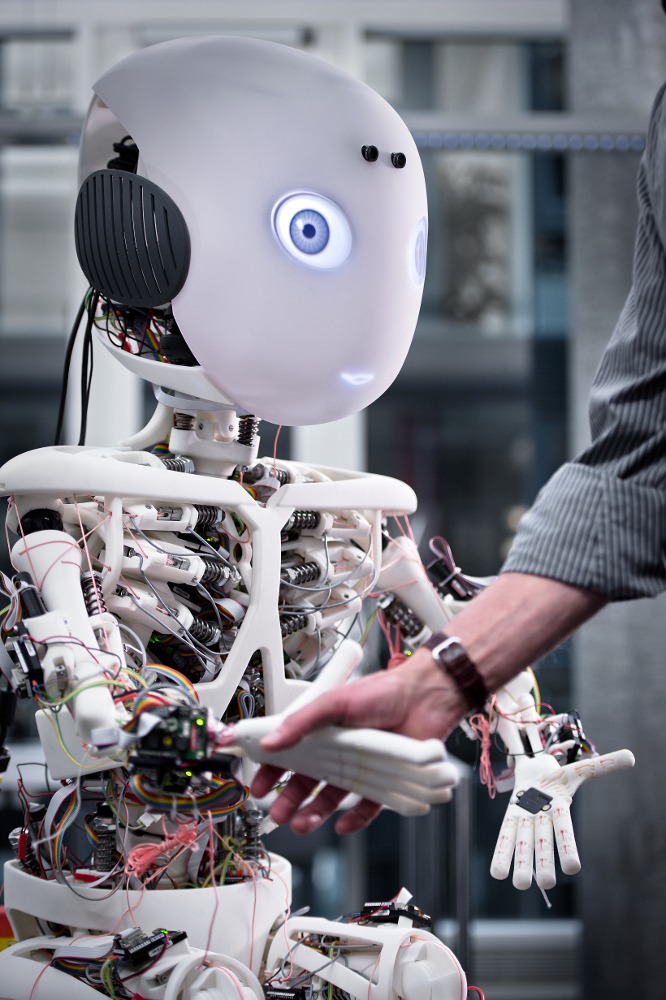}
	\caption{Roboy, a human-like, musculoskeletal robot with 28 degrees of freedom and 48 motors, to be controlled by brain-inspired systems. Photography: Adrian Baer}
	\label{fig:roboy}
\end{figure}
Neural models available for either PyNN or Nengo include diverse brain structures. 
In fact, the world's largest functional brain model, Spaun\cite{eliasmith2012}, is defined in Nengo. 
An embodied version of the model that can interact with the physical world as well as with humans would be an interesting test bed for human-robot interaction and cognitive science\cite{Krichmar2012}. 
A Spaun-like brain model combined with advanced musculoskeletal robots like Roboy\cite{Pfeifer2013} (Figure \ref{fig:roboy}) would herald a whole new era of robotic research. 
Our proof-of-concept system combining SpiNNaker and Myorobotics paves the way for exactly these kinds of endeavors, which we hope to stimulate with this article. 

A typical human cerebellum comprises about 100 billion neurons \cite{Andersen1992}, about as many as the rest of the brain. 
A realistic simulation of such a complex large scale system will rely not just on massive computing resources, it also requires a detailed and realistic environment to interact with. 
Therefore, real-time capable neuro-simulators in conjunction with robots will eventually become an essential tool of brain research. 
Practical and scalable systems like the one hereby presented thus enable an interaction between neuroscience and robotics which is mutual: 
Robots can help to advance neuroscience just as neuroscience helps us to create more natural robots. 

\section*{Acknowledgment}
We thank N. Luque for helpful discussions regarding cerebellar motor control, S. Temple and the SpiNNaker Manchester team for their invaluable hardware, software and support, and the Myorobotics team for providing us with robot parts. C.R. and J.C. acknowledge funding and support by the German Federal Ministry for Education and Research through the Bernstein Center for Computational Neuroscience Munich (01GQ1004A). S.J., R.H., A.K., F.R. and P.v.d.S. acknowledge funding from the European Union Seventh Framework Programme (FP7/2007-2013) under grant agreement no. 604102 (Human Brain Project), R.H. under grant agreement no.\ 288219 (Myorobotics). P.v.d.S. also acknowledges support from DLR. J.G and E.R. would like to acknowledge Spanish National Project NEUROPACT (TIN2013-47069-P). J.G. also acknowledges funding from the University of Granada and the European Union H2020 Framework Programme (H2020-MSCA-IF-2014) under grant agreement no. 653019 (CEREBSENSING).

\ifCLASSOPTIONcaptionsoff
  \newpage
\fi



\bibliographystyle{IEEEtran}
\bibliography{IEEEabrv,spice-paper-ram}

\end{document}